\documentclass{article}

\PassOptionsToPackage{dvipsnames}{xcolor}

\usepackage[final]{ewrl_2026}

\usepackage[utf8]{inputenc} 
\usepackage[T1]{fontenc}    
\usepackage{hyperref}       
\usepackage{url}            
\usepackage{booktabs}       
\usepackage{amsfonts}       
\usepackage{amssymb}        
\usepackage{nicefrac}       
\usepackage{microtype}      
\usepackage{graphicx}       
\usepackage{subcaption}     
\usepackage[dvipsnames]{xcolor}
\usepackage{colortbl}       
\usepackage{float}          
\usepackage{natbib}         
\usepackage{paralist}       

\usepackage[acronym]{glossaries}
\usepackage[inline]{enumitem}
\usepackage{fontawesome5}
\usepackage{xspace}
\usepackage[noabbrev,capitalize,nameinlink]{cleveref}

\glsdisablehyper 

\usepackage{pifont}
\usepackage{xspace}
\definecolor{green_d}{RGB}{46, 125, 50}
\definecolor{red_d}{RGB}{198, 40, 40}
\definecolor{orange_d}{RGB}{255, 193, 7}

    \newcommand{\yes}{\color{green_d}\ding{51}\xspace}
    \newcommand{\no}{\color{red_d}\ding{55}\xspace}
    \newcommand{\partialo}{\color{orange_d}$\blacktriangle$\xspace}

\title{Factoriax: A GPU-Accelerated Factorio-Style Simulator for Reinforcement Learning}

\author{%
  Mickey Beurskens\thanks{Corresponding author: \texttt{m.r.m.beurskens@tue.nl}}
  \qquad Tristan Tomilin
  \qquad Thiago D.~Sim\~{a}o \\
  Eindhoven University of Technology
}

\newacronym{rl}{RL}{Reinforcement Learning}
\newacronym{or}{OR}{Operations Research}
\newacronym{fle}{FLE}{Factorio Learning Environment}
\newcommand{\easyRocket}{\texttt{Easy Rocket}\xspace}

\usepackage[textwidth=15mm,
	backgroundcolor=white,
    tickmarkheight=2pt,
	textsize=tiny]{todonotes}

\definecolor{someblue}{HTML}{598BE7}
\colorlet{linkcolor}{someblue}
\hypersetup{colorlinks=true,urlcolor=linkcolor,citecolor=linkcolor,linkcolor=linkcolor}

\begin{document}
\maketitle
\setcounter{footnote}{0} 

\begin{abstract}
We introduce Factoriax, a GPU-accelerated factory-building simulator written in JAX.
In Factoriax, an agent must collect resources, build machines using those resources, and then automate the collection and crafting process by arranging machines on the map to build production pipelines. 
This paper discusses the structure of the Factoriax simulator and an initial benchmark called \easyRocket, in which an agent is tasked with building a resource-intensive machine called the Rocket in a limited number of game ticks to escape the planet.
We also publish results from a number of PPO-based training runs on \easyRocket.
Factoriax is built to be fast.
A 1-billion-step PPO training run, equivalent to 500,000 episodes, runs on \easyRocket in about 8 minutes on a single NVIDIA A100.
A standard laptop GPU can complete the same run in about 84 minutes. 
Our trained PPO agent learns to gather resources, craft machines from those resources, and place them on the map through a curriculum reward directly tied to a manually designed set of achievements.
After training, the agent does not place machines in a functional spatial configuration, failing to fully complete the benchmark, and leaving the challenge open for future attempts. 
\end{abstract}

\section{Introduction}
\begin{figure}[tbp]
  \centering
  \includegraphics[width=\linewidth,trim={10pt 0pt 10pt 0pt},clip]{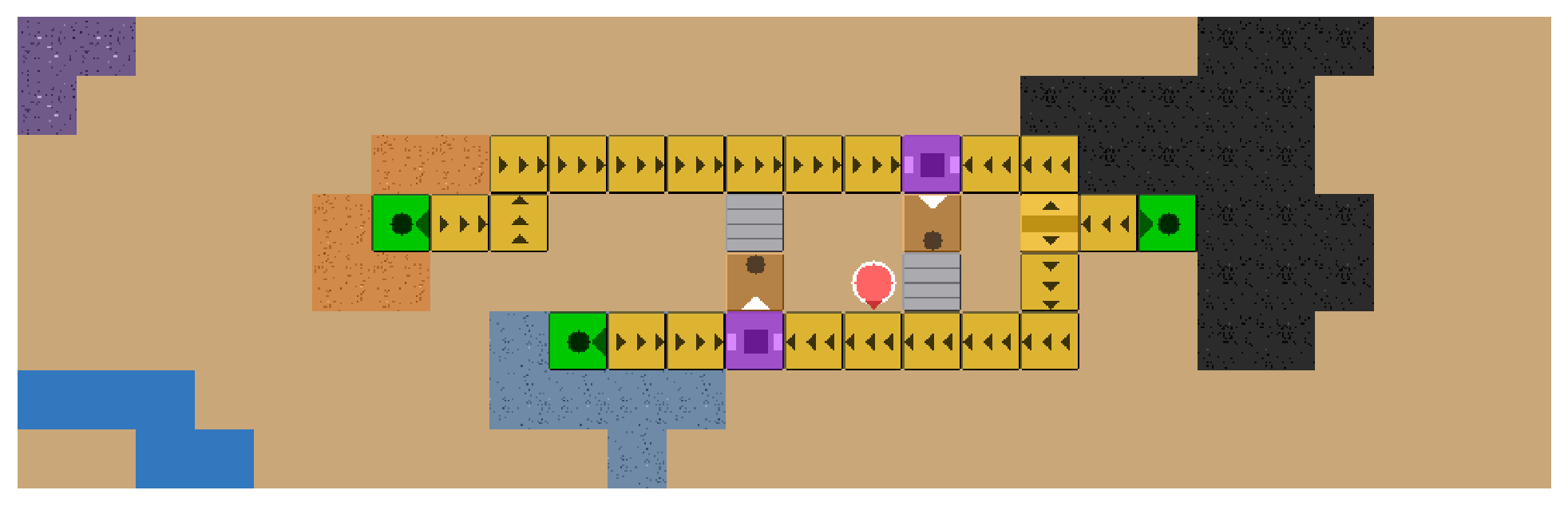}
  \caption{A typical early-game state in Factoriax. An agent (red dot) on a 2D grid mines raw materials using a miner (green), crafts machines, and arranges them so belts (yellow) carry the correct inputs to assemblers (purple). The assemblers create half-fabricates to speed up the production of more machines. A single coal miner on the right drops coal ore onto a belt, which is split between both lines through a belt splitter. Arms (brown) drop each assembler's output onto pallets (grey) acting as high-capacity item buffers.}
  \label{fig:title}
\end{figure}

We introduce Factoriax\footnote{\url{github.com/mickeybeurskens/factoriax}}, a GPU-accelerated factory building simulator written in JAX~\citep{jax2018github} for reinforcement learning (RL).
In Factoriax, an agent spawns in a grid world (\cref{fig:title}), where it must collect resources, build machines, and ultimately automate resource collection and processing by building and maintaining factories.
Factoriax enables research into factory layout design, often called facility layout problems~\citep{DBLP:journals/arc/BouramtaneKRBC24} in operations research (OR), with RL.
Approaches to solve factory layout design tend to treat the problem as disembodied, with the inputs being some sort of matrix or graph and the output being a permutation of the factory layout. 
Studying embodied agents in a dynamically built factory setting such as Factoriax is less common.
The order of construction becomes more important for an agent moving in an embodied way through a simulator, as it might create or obstruct its own routing.
Concerns like reach and walking distance become apparent as well. 
Embodiment thus adds an additional spatial element to the planning, execution, and maintenance of factory layout design.

Factoriax is inspired by the game mechanics featured in Factorio\footnote{\url{factorio.com}}.
The Factorio Learning Environment (FLE)~\citep{Hopkins2025Factorio} uses the full Factorio game as a language model benchmark.
However, the game is optimized for human gameplay and runs at a limited speed of 60 frames per second, making FLE impractical for RL.
Although less faithful to the original game than FLE, Factoriax preserves the core challenges of long-horizon planning, layout design, and agent embodiment that make the game interesting for RL research. 
Moreover, by leveraging JAX to load both the model and the environment onto the GPU,  Factoriax reduces communication overhead, resulting in significant improvements in computational throughput.

In addition to the simulator behind Factoriax and its key design ideas, this paper introduces \easyRocket, the first benchmark built in Factoriax.
A PPO-based agent shows that \easyRocket is partly solved.
Solving the full benchmark is still an open problem.
Finally, this work includes a computational performance analysis of \easyRocket and the PPO training run.
It shows that a 1-billion-step PPO training run on the included \easyRocket benchmark completes in about 8 minutes of wall-clock time on a single NVIDIA A100, and about 84 minutes on a standard laptop GPU. 

\section{Related Work}

Factoriax combines \textit{GPU acceleration} and a focus on \textit{embodied} agents and factory \textit{layout design}. This section discusses related benchmarks along these lines. \cref{tab:benchmark_comparison} provides an overview of the benchmarks and their relationship to Factoriax.

\paragraph{GPU acceleration} 
Some simulators are built to run on the GPU, which are then used to create different experimental settings and benchmarks.
An example in 3D physics is MuJoCo MJX~\citep{mujocoMuJoCoMJX}, the GPU-accelerated version of the simulator that powers the robotics environments in the Gymnasium~\citep{Towers2025Gymnasium}, the accelerated continuous control benchmark Brax~\citep{DBLP:conf/nips/FreemanFRGMB21}, and a safe reinforcement learning benchmark  CRAX~\citep{Tomilin2026CRAX}.
Other simulators include IsaacLab~\citep{mittal2025isaaclab} for 3D physics and Kinetix~\citep{Matthews2025kinetix} for 2D physics. 
Similarly, Factoriax uses JAX to leverage GPU acceleration to speed up simulation.

Examples in other fields using GPU acceleration include Gymnax~\citep{gymnax2022github}, which implements JAX versions of classical problems like CartPole, MountainCar, and the Atari environment.
Pgx~\citep{Koyamada2023Pgx} implements popular board games like chess and Go.
JaxMARL~\citep{DBLP:conf/nips/RutherfordEG0LI24} implements a host of multi-agent benchmarks and algorithms in JAX.
MEAL~\citep{tomilin2026meal} extends multi-agent benchmarking specifically for continual RL. 
SocialJax~\citep{guo2025socialjax} is more geared towards studying sequential social dilemmas.
Finally, Jumanji~\citep{DBLP:conf/iclr/BonnetLBSADCMTK24} focuses on games like Tetris, Snake, Maze, and the Rubik's Cube.

\paragraph{Embodiment}
Embodiment broadly refers to a benchmark in which the agent is located in the environment and has to navigate and move around to complete its goals.
The robotics benchmarks discussed earlier belong to this category. 
Many video games are also used as embodied agent benchmarks. 
Minecraft has been used as a crafting-centric embodied benchmark through MineRL~\citep{Guss2019MineRL}.
Doom has been adapted into benchmarks geared to specific reinforcement learning settings, such as COOM~\citep{Tomilin2023COOM} for continual learning, and HASARD~\citep{Tomilin2025Hasard} for safe RL.
Another approach to simulate embodied agents is using a grid world.
Factoriax uses this approach, as do other benchmarks.
Craftax~\citep{DBLP:conf/icml/MatthewsBESJCF24}, inspired by the NetHack Learning Environment~\citep{Kuttler2020NetHack}, features an embodied agent in a grid-world RPG setting with procedural level generation, dungeon exploration, item crafting, and persistent block placement. 
XLand MiniGrid~\citep{DBLP:conf/nips/NikulinKZASK24} is an accelerated benchmark aimed at studying meta-reinforcement learning with a dynamically configurable grid world.

\begin{table}[t]
  \centering
  \caption{Comparing Factoriax to other simulators and benchmarks. Comparisons assess whether a benchmark is \textit{GPU-accelerated}, whether the agent in the benchmark is \textit{embodied} and has to navigate through the environment, and whether the benchmark tests the agent's capacity to create interconnected factory \textit{layout designs}. {\partialo} indicates partial fit.
  }
  \label{tab:benchmark_comparison}
  \resizebox{\linewidth}{!}{\newlength{\gapheight}
\setlength{\gapheight}{1.1em}
\begin{tabular}{@{}lccclr@{}}
    \toprule
     & \textbf{GPU} & \textbf{Embodiment} & \textbf{Layout}  & \textbf{Details} & \textbf{Reference} \\
    \midrule
    SafeOR-Gym       & \no  & \no & \partialo & Operations research   & \citep{Ramanujam2026safeORgym} \\[\gapheight]
    Gym / Gymnasium  & \no  & \yes & \no  & Standard API, classic control          & \citep{Towers2025Gymnasium} \\
    NetHack          & \no  & \yes & \no  & Roguelike (procgen)                    & \citep{Kuttler2020NetHack} \\
    COOM             & \no  & \yes & \no  & Doom (continual RL)                    & \citep{Tomilin2023COOM} \\
    HASARD           & \no  & \yes & \no  & Doom (safe RL)                         & \citep{Tomilin2025Hasard} \\
    MineRL           & \no  & \yes & \partialo & Minecraft building                & \citep{Guss2019MineRL} \\[\gapheight]
    FLE              & \no  & \yes & \yes & Factorio for LLM agents                & \citep{Hopkins2025Factorio} \\[\gapheight]
    Pgx              & \yes & \no  & \no  & Board games (chess, go)                & \citep{Koyamada2023Pgx} \\
    Jumanji          & \yes & \no & \no       & Combinatorial OR     & \citep{DBLP:conf/iclr/BonnetLBSADCMTK24} \\[\gapheight]
    MuJoCo XLA~(MJX)       & \yes & \yes & \no  & 3D physics engine                      & \citep{mujocoMuJoCoMJX} \\
    IsaacLab         & \yes & \yes & \no  & 3D robotics simulation                 & \citep{mittal2025isaaclab} \\
    Kinetix          & \yes & \yes & \no  & 2D physics engine                      & \citep{Matthews2025kinetix} \\
    CRAX             & \yes & \yes & \no  & Safe RL (MuJoCo-based)                 & \citep{Tomilin2026CRAX} \\
    Brax             & \yes & \yes & \no & 3D continuous control                   & \citep{DBLP:conf/nips/FreemanFRGMB21} \\
    SocialJax        & \yes & \yes & \no & Social dilemmas                         & \citep{guo2025socialjax} \\
    XLand-MiniGrid   & \yes & \yes & \no & Meta reinforcement learning             & \citep{DBLP:conf/nips/NikulinKZASK24} \\
    Gymnax           & \yes & \yes & \no & Collection of environments              & \citep{gymnax2022github} \\
    JaxMARL          & \yes & \yes & \no & Multi-agent                             & \citep{DBLP:conf/nips/RutherfordEG0LI24} \\
    MEAL             & \yes & \yes & \no  & Continual multi-agent                  & \citep{tomilin2026meal} \\
    Craftax          & \yes & \yes & \partialo & Open-ended survival               & \citep{DBLP:conf/icml/MatthewsBESJCF24} \\[\gapheight]
    \textbf{Factoriax} & \yes & \yes & \yes & Factory building / coupled OR & This paper\\
    \bottomrule
\end{tabular}%
}
\end{table}

\paragraph{Layout design}
SafeOR-Gym~\citep{Ramanujam2026safeORgym} decomposes OR problems into nine operations-research tasks as separate constrained environments for safe RL.
Factoriax differentiates itself by focusing on a GPU-accelerated combination of embodiment and factory layout design.
The Factorio Learning Environment~(FLE)~\citep{Hopkins2025Factorio} is the only other benchmark combining embodiment and layout design. 
FLE exposes a software interface for large language models to interact directly with the full Factorio game.
The full game of Factorio has much richer game mechanics, but is not GPU-accelerated.
FLE runs at Factorio's native speed of roughly 60 game ticks per second, which is too slow to train and evaluate RL algorithms within time frames practical for most researchers in the RL community.
While Factoriax does not have the same fidelity to the original game as FLE, it preserves the core challenges of long-horizon planning, layout design, and agent embodiment. 

\section{Factoriax Simulator}
\label{sec:simulator}

\subsection{Mechanics and Gameplay}
\paragraph{Game Loop}
\cref{fig:title} shows the central gameplay loop in Factoriax.
At the start of a run (episode), the agent spawns in a bounded grid-world where each location may contain a number of resources, a machine, or the agent's character.
Each action the agent takes progresses the entire world by a single game tick. 
Resources can be mined and subsequently hand-crafted by the agent into half-fabricates and machines.
Machines can be placed to automate parts of the crafting chain, allowing the agent to expand the factory setup while production continues without direct agent interaction.

\paragraph{Termination}
Runs can have different win conditions depending on the scenario.
For example:
\begin{itemize}
    \item ``Produce an expensive machine as fast as possible.'' 
    \item ``Produce as many items of a specific type as you can in a limited amount of time.''
    \item ``Balance the throughput of an existing factory.''
    \item ``Triage a resource depletion issue by re-routing the factory to restart production.''
\end{itemize}
The environment resets when the win condition is achieved.
Resetting might generate a new randomized map, or pick from a set of hand-designed levels, depending on the setup.

\begin{figure}[t]
  \centering
  \includegraphics[width=1\linewidth]{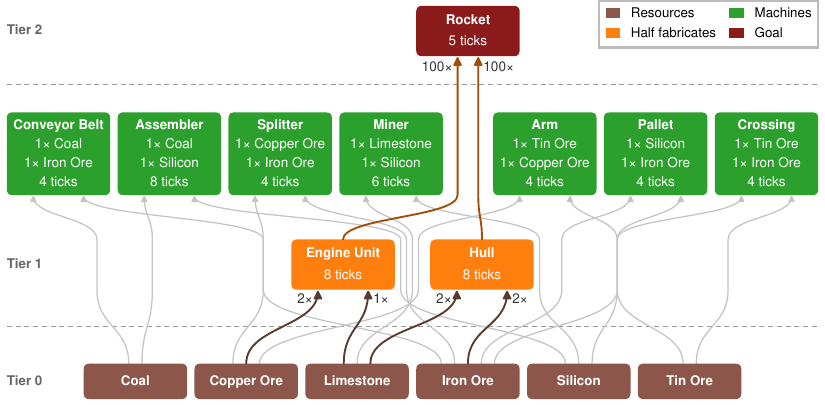}
  \caption{A custom recipe book designed for the \easyRocket task (\cref{sec:benchmark}). Raw ores feed every intermediate. The rocket itself requires 100 hulls and 100 engine units (highlighted edges). This requires a working production line rather than hand assembly to finish within 2000 ticks. \cref{app:items} shows the full standard recipe book.}
  \label{fig:recipe_dag}
\end{figure}

\paragraph{Crafting} 
Crafting transforms items according to the so-called recipes.
Each recipe combines two item types, which can be raw resources, half-fabricates, or machines.
The number of items required as input and collected as output can be configured differently depending on the scenario.
A scenario can use the default recipe book or create a custom recipe book containing different item combinations. 
\cref{fig:recipe_dag} shows a customized recipe book and
\cref{fig:full_recipe_dag} shows the default version.
Each input combination can only occur once per recipe book, which is checked during the initialization of a scenario.
This makes each recipe book a directed acyclic graph (DAG).

\paragraph{Machines} Machines can be placed on the map while raw resources and half-fabricates cannot.
\cref{tab:machines} provides an overview of the most important machines.
Miners play a crucial role by extracting some amount of raw ore from the ground during each game tick and placing the ore on an available conveyor belt in the direction it faces.
Conveyors, splitters, and crossings carry a configurable amount of items across them each game tick. Assemblers can craft recipes using the items on two input conveyors. Arms can pick items out of one machine inventory and put them in another, adding to the available routing options in the game. Pallets act as high-capacity item buffers. Finally, the rocket is a special item that can be used as a capstone or finishing condition. Most machines have an internal item buffer that can store a limited number of items. All these machines form the basic building blocks for a factory in Factoriax.

\paragraph{Items} The agent has an internal inventory with space for a fixed number of each item, similar to the player carrying a specific ``pouch'' for every item of the game.
\cref{app:items} describes all the ores, half-fabricates, and machines currently implemented in Factoriax. 

\begin{table}[tbp]
  \centering
  \caption{Subset of the placeable machines in Factoriax. Icons are the in-simulator sprites used by the renderer. \cref{app:items} gives an overview of the machines' crafting recipes.}
  \label{tab:machines}
  {
\begin{tabular}{@{}c l l p{0.30\linewidth} p{0.30\linewidth}@{}}
\toprule
 & Name & Role & Per-tick action & Key property \\
\midrule
\raisebox{-0.32\height}{\includegraphics[width=18pt]{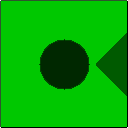}} & Miner & Producer & Pulls ore from the tile beneath into its output buffer & Needs an ore patch beneath; depletes the patch \\
\raisebox{-0.32\height}{\includegraphics[width=18pt]{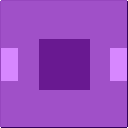}} & Assembler & Producer & Pulls two inputs, advances a recipe timer, emits the output & 2-input $\to$ 1-output; recipe set by inserted items \\
\raisebox{-0.32\height}{\includegraphics[width=18pt]{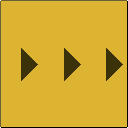}} & Conveyor belt & Logistics & Shifts contents one tile forward along its facing direction & Two parallel lanes (A and B) \\
\raisebox{-0.32\height}{\includegraphics[width=18pt]{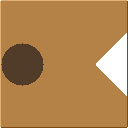}} & Arm & Logistics & Picks one item from the tile it faces, deposits it on the opposite tile & One item transferred per tick \\
\raisebox{-0.32\height}{\includegraphics[width=18pt]{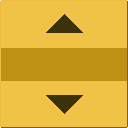}} & Splitter & Logistics & Alternates incoming belt items between two outputs & Balances two downstream consumers from one supply \\
\raisebox{-0.32\height}{\includegraphics[width=18pt]{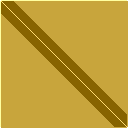}} & Crossing & Logistics & Two belt lanes that cross perpendicularly without merging & No item transfer between the crossing lanes \\
\raisebox{-0.32\height}{\includegraphics[width=18pt]{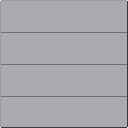}} & Pallet & Storage & Passive single-slot storage & Items picked up by arms or by the player \\
\raisebox{-0.32\height}{\includegraphics[width=18pt]{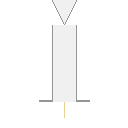}} & Rocket & Capstone & No per-tick action; placement is the event & Scenarios may treat placement as a capstone signal \\
\bottomrule
\end{tabular}
}

\end{table}

\subsection{RL Agent Interface}
\label{sec:components}

\paragraph{Observations}
Factoriax has several options for agent observations, mainly divided into two axes.
First, an agent might use \textit{x-ray} to see the internal state of every machine and resource in its observational range. 
Otherwise, the observation is \textit{superficial}, and the agent can only see detailed information about the machine it is facing. 
This setting was added to support partial-observable scenarios later on.
The second axis pertains to global versus local observations. 
The global observation encodes the world in absolute coordinates and shows every tile to the agent. 
Local observations are encoded relatively and show tiles up to a configured radius. 
All four variants emit flattened vectors and a per-player scalar block for pose, inventory state, and a per-recipe flag that indicates whether a recipe can be manually crafted by the agent.
\cref{app:observation} catalogues every observation channel.

\paragraph{Actions}
Agents can choose actions based on one of five categories: \texttt{move}, \texttt{interact}, \texttt{place}, \texttt{craft}, and \texttt{deposit}.
\texttt{Move} includes lateral translation and rotations, as well as the \texttt{NOOP} action, meaning the agent skips acting for a turn and progresses time by a single tick.
\texttt{Interact} covers mining the currently faced tile, picking up a machine, withdrawing items from a machine item buffer, and machine rotation on the map. 
The other three categories are parametric.
\texttt{Place} has one action for every placeable machine.
A machine can only be placed if it is present in the agent's inventory. 
\texttt{Craft} has one action for every half-fabricate and machine and only triggers if all the required base materials are present in the inventory.
\texttt{Deposit} has an action for every item, including raw resources. 
The agent can choose to deposit any item into the item buffer of a machine it faces.
Factoriax includes an \texttt{ActionMaskWrapper} that can be used to hide actions from the agent. 
A masked action defaults to \texttt{NOOP}.
\cref{app:action} describes the full set of available actions.

\paragraph{Rewards}
Factoriax contains a host of reward functions.
The implementation splits into two types.
One type of reward is granted based only on the current state. 
Examples include \texttt{mining\_reward} and \texttt{sparse\_mining\_reward}.
No other information is needed.
Other rewards are constructed using not only the current state, but also whether that state has already occurred.
If it has, no reward is given.
Factoriax implements this with an \textit{achievement system}.
Each achievement is tied to a particular feature of world state. 
A particular benchmark implementation decides the conditions that trigger an achievement. 
The achievement is then recorded and passed along so that it is remembered throughout the episode.

\subsection{Research Tools}

Factoriax aims to facilitate research by providing an analysis library, a human-playable version of the game, and a level editor to support manual design of new levels and benchmarks.
The analysis module contains standard plotting tools, a standard episode recording format, and ways to create visual snapshots and video files of the game and the internal agent inventory.
The human-playable version of the game loads a randomized map in ``free play'' mode.
It can be played with a keyboard or a controller.
The level editor has a feature to manually draw ore patches and place machines.
Levels can be loaded in the human-playable interface directly from the editor for testing and tweaking.

\section{Easy Rocket Benchmark}
\label{sec:benchmark}

In this section, we introduce \easyRocket, the first benchmark in Factoriax.
The goal in \easyRocket is to produce a machine called the Rocket and place it on the map as fast as possible.
The benchmark features a simplified recipe book, shown in \cref{fig:recipe_dag}, with at most two production steps, called tiers, to get to the final product. The Rocket needs only 100 \texttt{engine units} and 100 \texttt{hulls}; however, crafting such resources directly is infeasible in the available 2000 ticks.
Therefore, the agent must build a factory to gain efficiency and complete the challenge on time.

The map is a 16-by-16 grid, with the agent always spawning at the center, as shown in \cref{fig:reset_grid}, and
each resource spawns in a 2-by-2 patch at a random location.
To support reproducibility, this randomization produces the same map given a fixed seed.
Other benchmark designs could have a small amount of resources per tile, forcing the agent to triage resource depletion problems during the challenge.
This is deliberately not the case in the \easyRocket benchmark.
One ore tile contains enough resources to make all the necessary parts to complete the benchmark.
The agent only needs to learn which ore types are relevant to build the components it needs.

\easyRocket runs on a \textit{superficial global} observation vector as described in \cref{sec:components}.  
The action space is unmasked.
The reward is manually designed to form a curriculum based on achievements.
When an achievement is attained, a reward is triggered.
The achievements can be divided into phases. 
The \textit{bootstrapping} phase focuses on basic navigation and resource collection.
The \textit{miners} phase automates resource extraction.
The \textit{storage} phase rewards forming the first connection between machines in the form of connecting miners to storage crates.
Finally, the \textit{hull}, \textit{engine}, and \textit{rocket production} phases all reward the production of the necessary half-fabricates and the rocket.   
Crucially, there are no achievements for connecting all of the machines in a particular way.
Achieving victory under these fairly sparse conditions is the core challenge of the \easyRocket benchmark.
\cref{app:easy_rocket} shows a breakdown of the achievements and associated trigger conditions.

\begin{figure}[t]
  \centering
  \begin{subfigure}[b]{0.55\linewidth}
    \centering
    \includegraphics[width=\linewidth]{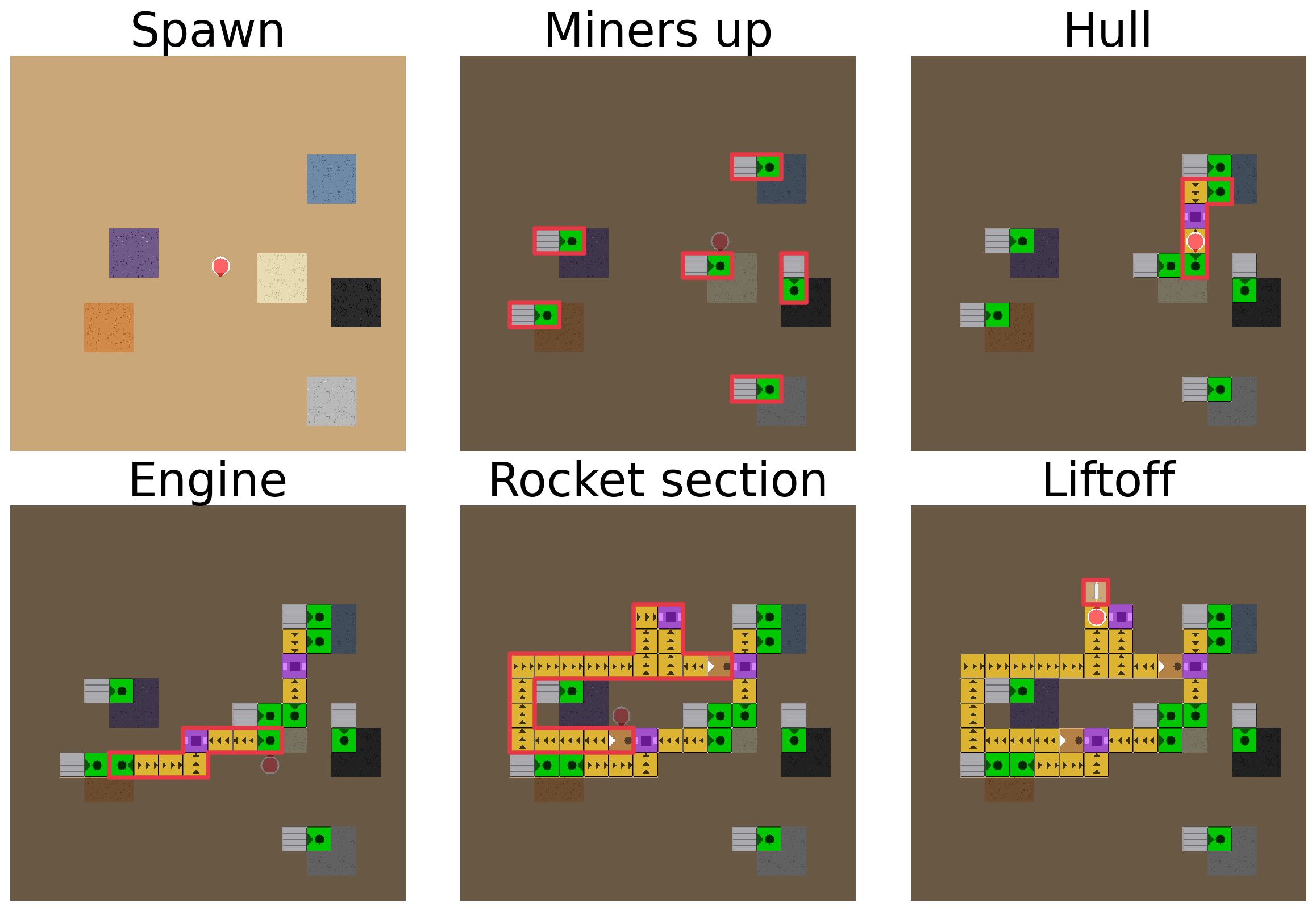}
    \caption{}
    \label{fig:build_progression}
  \end{subfigure}\hfill
  \begin{subfigure}[b]{0.43\linewidth}
    \centering
    \includegraphics[width=\linewidth]{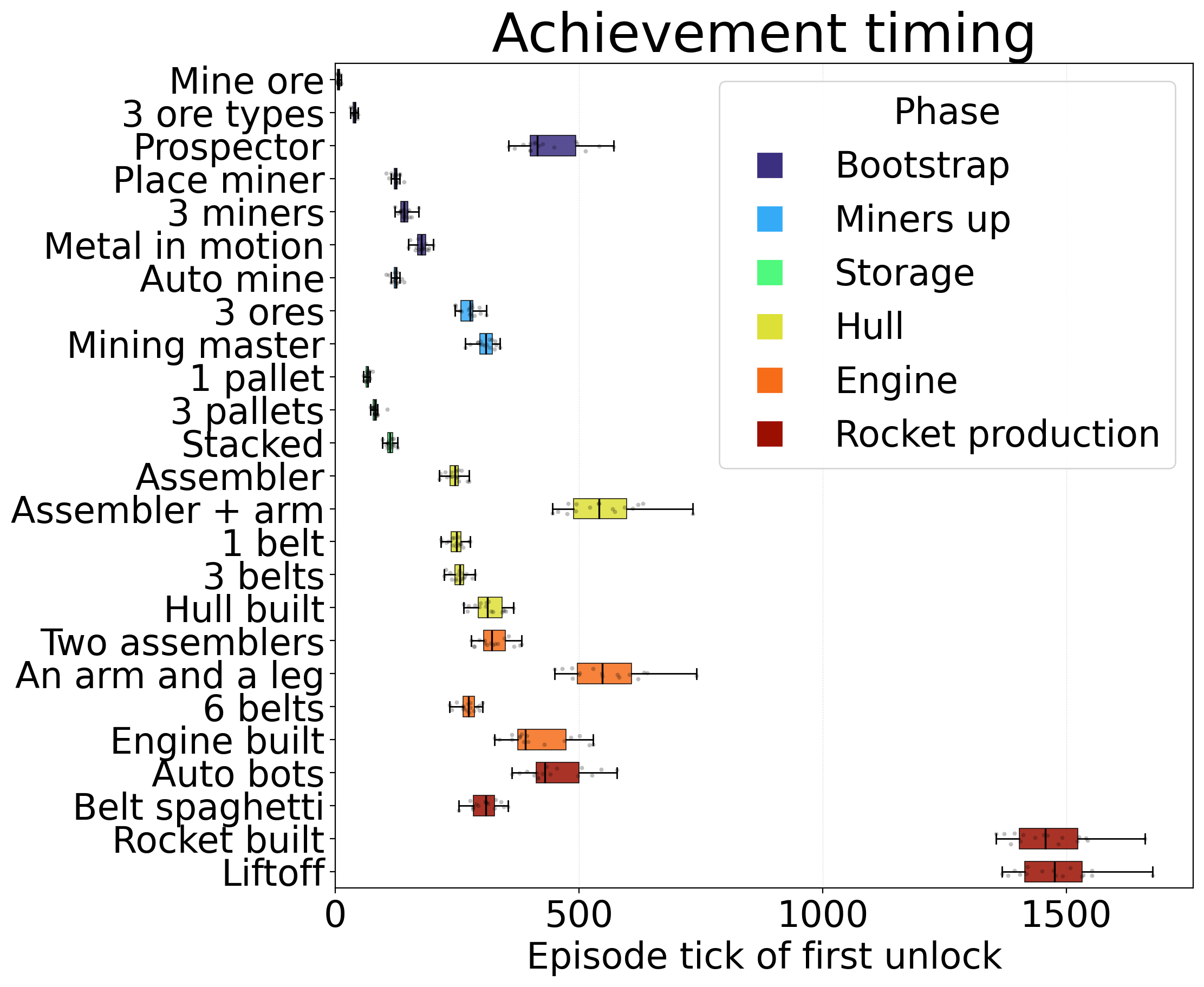}
    \caption{}
    \label{fig:scripted_timeline}
  \end{subfigure}
  \caption{Scripted-baseline agent. \textbf{(a)} Cumulative factory at the end of each phase on seed 42. \textbf{(b)} First-unlock tick distribution across 16 random map seeds. The scripted baseline reaches all twenty-five achievements on every seed within the 2000-tick budget.}
  \label{fig:scripted_baseline}
\end{figure}
\subsection{Scripted Baseline}

We use a scripted agent to validate the benchmark's feasibility, tune its parameters, and establish a baseline for comparison. 
We tested the scripted agent on 16 different map seeds, and it solved all of them.
The scripted agent's strategy is fixed.
Since each miner can mine three ores at a time by default, while the agent can only mine one ore per tick by hand, the scripted agent first places miners on every ore patch.
Next, the agent starts crafting machines for the hull, engine, and rocket section of the factory, returning to the miners for more base material between phases. 
Eventually the factory produces a Rocket, which is finally placed to complete the challenge. 
\cref{fig:build_progression} shows six snapshots from seed 42 corresponding to the described phases.
\cref{fig:scripted_timeline} indicates at which game tick achievements are completed across the 16 testing seeds. 
The strategy discussed in this section can be treated as a rough upper-bound performance estimate for the benchmark, although more efficient ways to solve it likely exist.

\begin{figure}[t]
  \centering
  \includegraphics[width=\linewidth]{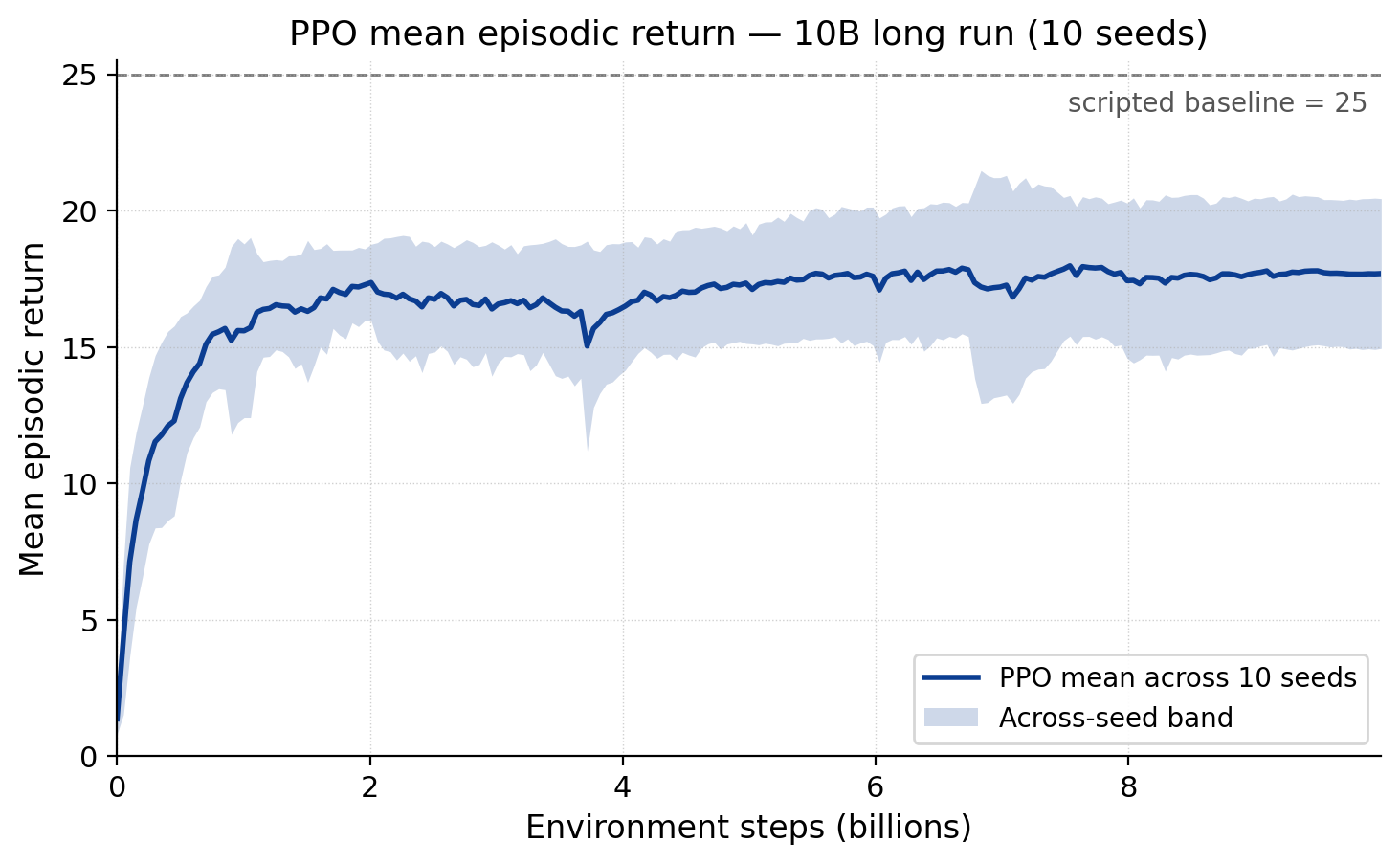}
  \caption{Mean episodic return on \easyRocket over training, across ten seeds at learning rate $10^{-4}$ and entropy coefficient $0.01$. The solid blue line shows the mean across seeds and the shaded area the standard deviation. The dashed reference at 25 is the scripted-baseline ceiling, or in other words, a fully successful run. Return saturates around 1.5 billion steps at a mean of $17.7$ and is flat thereafter.}
  \label{fig:ppo_returns}
\end{figure}
\section{Experiments}
\label{sec:experiments}

This section discusses the experimental setup for a PPO training run on the \easyRocket benchmark, as well as performance results in terms of episodic return and computational throughput. We train a PPO~\citep{DBLP:journals/corr/SchulmanWDRK17} agent on the \easyRocket benchmark on a total of ten independent seeds at learning rate $10^{-4}$ and entropy coefficient $0.01$.
Each agent is trained for 10 billion steps, totaling 100 billion steps.
The neural network is a small MLP architecture with two hidden layers of 256 neurons each. \cref{app:reproducibility} provides further details on the setup.

\subsection{PPO on Easy Rocket}

\begin{figure}[t]
  \centering
  \includegraphics[width=\linewidth]{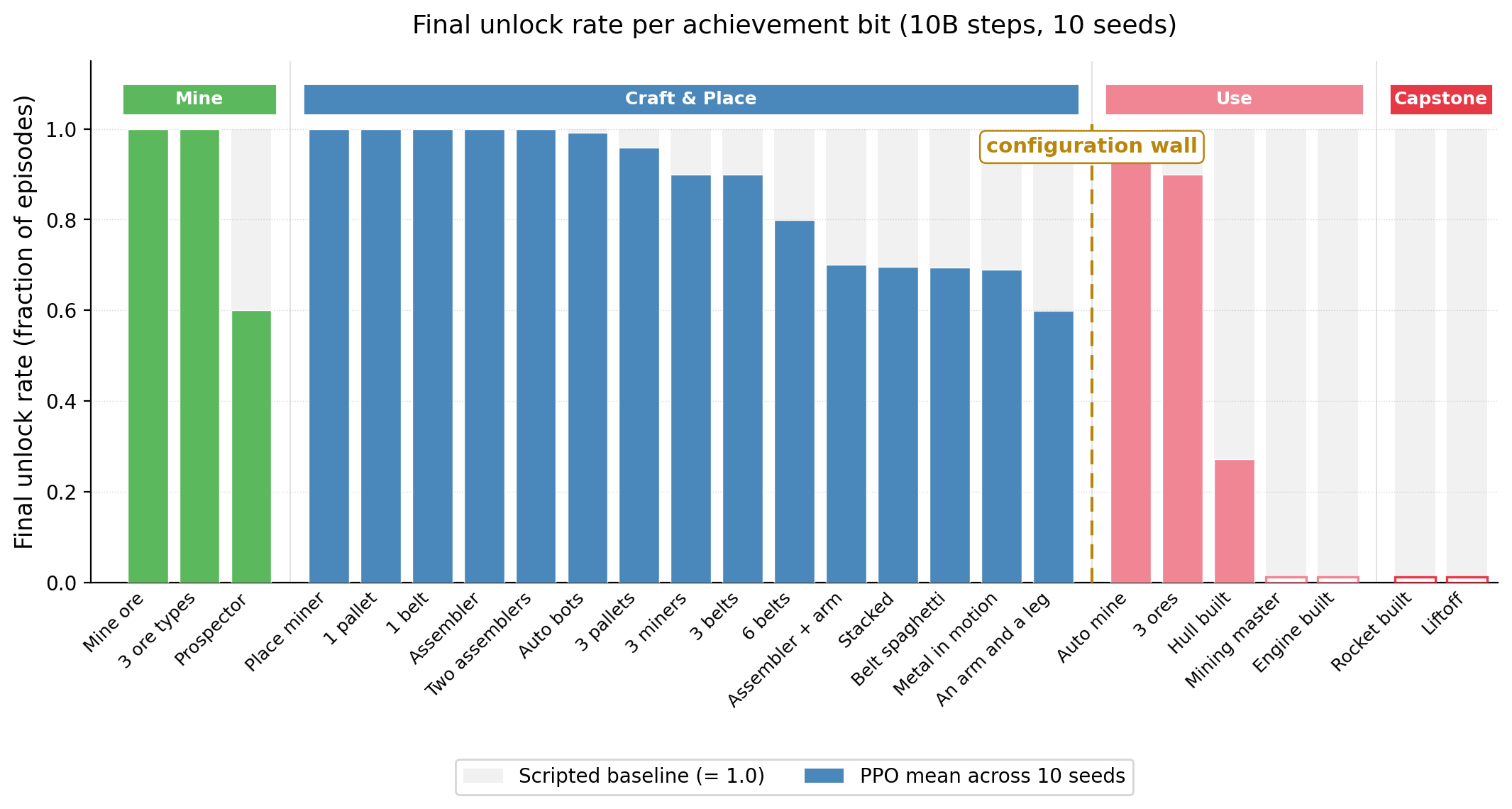}
  \caption{Per-achievement unlock rate at 10 billion steps across ten PPO seeds.}
  \label{fig:achievement_buckets}
\end{figure}

\cref{fig:ppo_returns} shows the training curve for the PPO training run. 
No run ever manages to complete the full benchmark. 
Mean episodic return saturates at $17.7 \pm 2.75$ of 25 possible achievements around 1.5 billion steps and is flat thereafter.
\cref{fig:achievement_buckets} shows the unlock rate of achievements divided into a number of phases.
The agent struggles to learn achievements related to the latter part of the \textit{use} phase, namely ``hull built'', ``engine built'', and ``mining master''.

Both ``hull built'' and ``engine built'' require completing two tiers of construction.
The difference between hull and engine production rates might have to do with the recipe book for \easyRocket, shown in \cref{fig:recipe_dag}.
Agents are rewarded for crafting miners and conveyor belts, which indirectly reward gathering iron ore, coal, limestone, and silicon. 
Since both iron ore and limestone are needed for hull production, the probability that the agent accidentally discovers how to craft a hull is quite high.
Compare this to the engine unit, which requires limestone and copper ore.
The only other device that uses copper ore is the arm, which is not connected to an achievement. 
The agent has no incentive to collect copper and is thus less likely to stumble upon unlocking the engine unit achievement.

In similar fashion, the agent never learns to complete ``mining master'', which requires at least one miner placed on every ore type. 
The agent is naturally incentivized to gather four different ore types to complete all the crafting achievements.
However, collecting tin or copper for crafting is never directly or indirectly rewarded.
To cross the gap and complete ``mining master'', an agent needs to collect materials, craft enough miners, walk to the remaining ore patches it never otherwise uses, and place the miners. 
This hardly ever happens by accident during exploration, which is why the agent never learns to complete the achievement.

The common theme between both failures to learn is that the curriculum does not expose the agent enough to the discovery of the hard-to-complete achievements.
In any case, the agent never learns to build even a partially operational factory, instead stopping at learning to craft and place machines in particular locations of interest.
The \easyRocket benchmark with the current reward structure thus remains an open problem.

\subsection{Throughput}
\label{sec:throughput}

This section describes a throughput analysis on an NVIDIA A100, referred to as the \emph{server GPU}, and an NVIDIA GeForce RTX 3050 6GB \emph{laptop GPU}.
We compare the throughput performance of raw environment runs and PPO training runs.
The PPO training run loads Factoriax into GPU memory alongside the training code for PPO, which makes it suitable for checking actual training performance.
Raw environment performance is checked by running the environments with random action selection. 
It is mainly useful to get a sense of the impact that different environment settings have on the throughput. 
The throughput for \easyRocket is mainly determined by three parameters: training batch size, entity limit, and map size.

\begin{figure}[t]
  \centering
  \includegraphics[width=\linewidth]{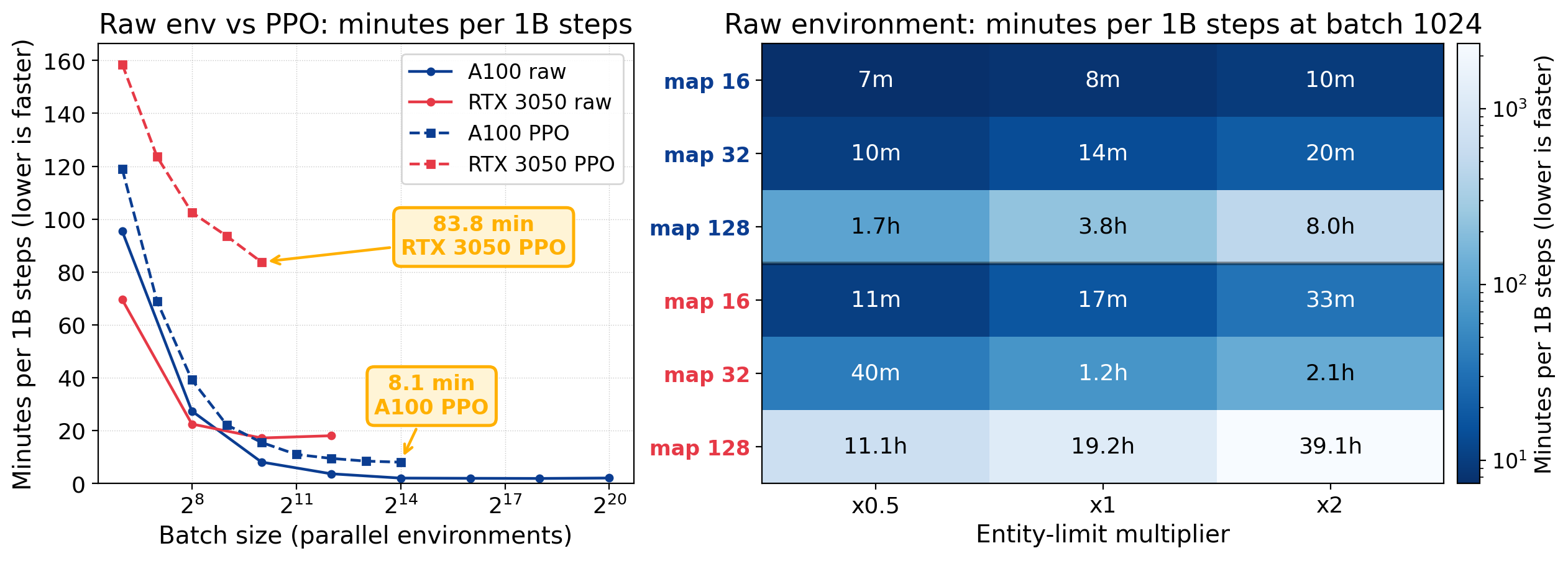}
  \caption{Left: A plot of wall-clock time versus the batch size on the GPU. Right: Impact on wall-clock time for map size and entity limit, where blue indicates the server GPU and red the laptop GPU.}
  \label{fig:throughput}
\end{figure}

\paragraph{Batch Size}
Batch size indicates the number of Factoriax environments that can run in parallel on the GPU.
The left image in \cref{fig:throughput} shows how batch size influences throughput. 
Batch size is mainly limited by the Video Random Access Memory (VRAM) of the GPU. 
The laptop GPU used here has much lower VRAM than the server GPU and can thus serve fewer concurrent environments at a time.
At some point, reading and writing to the GPU becomes the bottleneck for training, and the rate of performance gain drops off.
A consequence of a large batch size is that a larger amount of training happens in parallel.
If the total number of episodes in the training is lower than the number of episodes represented in the training set, the model can only update a single time.
Therefore, the batch size also needs to be tuned such that enough training updates can occur during training.
The wall-clock training time of a 1-billion-step PPO training run, or approximately 500,000 episodes, on the laptop GPU is approximately 84 minutes.
On the server GPU, the same run takes about 8 minutes of wall-clock time.

\paragraph{Entity Limit and Map Size}

Factoriax optimizes performance by defining a hard limit on the number of machine entities on a map.
This generally works well since the number of placed machines is usually much lower than the number of tiles in a Factoriax environment for all but the smallest maps.
As can be seen on the right side of \cref{fig:throughput}, a change in the entity limit can have a large impact on wall-clock time for raw environment throughput. 
For both the laptop GPU and the server GPU, varying between half the standard entity limit and double the standard entity limit results in around 40 percent increased wall-clock time for a run on the lower end of the map sizes.
This can go as high as a difference of almost five times the wall-clock time on the server GPU with map size 128 by 128. 
Map size itself also has a large impact on performance across the board for both the laptop GPU and the server GPU.

\section{Conclusion}
\label{sec:conclusion}

We introduced Factoriax as a new factory-building simulator for RL, and \easyRocket, a curriculum reward benchmark built with Factoriax.
Factoriax uniquely combines GPU acceleration, embodied agents, and factory layout design in a single simulator.
The purpose of an agent in a Factoriax environment is to gather resources and build a factory. 
The PPO agent partially solved the \easyRocket benchmark, but could not complete the full challenge. 
Specifically, it could not learn how to connect machines to build a functional factory, although it did learn to craft and place machines on the map and complete a large part of the associated achievements.
This means \easyRocket remains an open problem.
\easyRocket PPO training runs with 1 billion steps can be completed in 84 minutes on a laptop GPU and 8 minutes on an NVIDIA A100.
We bring this Factoriax to EWRL hoping to find out which future developments matter most to the community, and we look forward to people attempting the \easyRocket challenge themselves.


\newpage
\bibliographystyle{plainnat}
\bibliography{references}

\clearpage
\crefalias{section}{appendix}
\appendix

\section{AI Use Disclaimer}
AI was used to generate the initial version of this paper. It was subsequently rewritten entirely by hand because the resulting writing was mediocre and confusing at best.
Hopefully the rewrite has improved upon that.
AI was useful to check for grammatical mistakes and perform broad scans on related works.

The initial version of the Factoriax code was generated mainly by AI as an experiment in the possibilities of AI-driven coding.
The process was anchored in test-driven development.
Every feature included in the simulator had an associated test to ensure correct implementation. 
Despite this effort and others, the code turned out ugly and full of ``code smell''.
Therefore, the authors ended up also performing significant manual inspection and refactoring on the code.
During this process, many bugs surfaced that would otherwise have gone undetected, even after performing multiple automated bug scans using the same models. 
AI was fruitfully used to perform large scans on the codebase to search for implementation details, understand JAX, create plots, and create prototypes of both the JAX implementations and UI elements for comparison. 

We hope these short observations will help other researchers decide how to apply AI to their own research projects.

\newpage
\section{Item Reference}
\label{app:items}

\begin{figure}[H]
  \centering
  \includegraphics[width=0.95\linewidth]{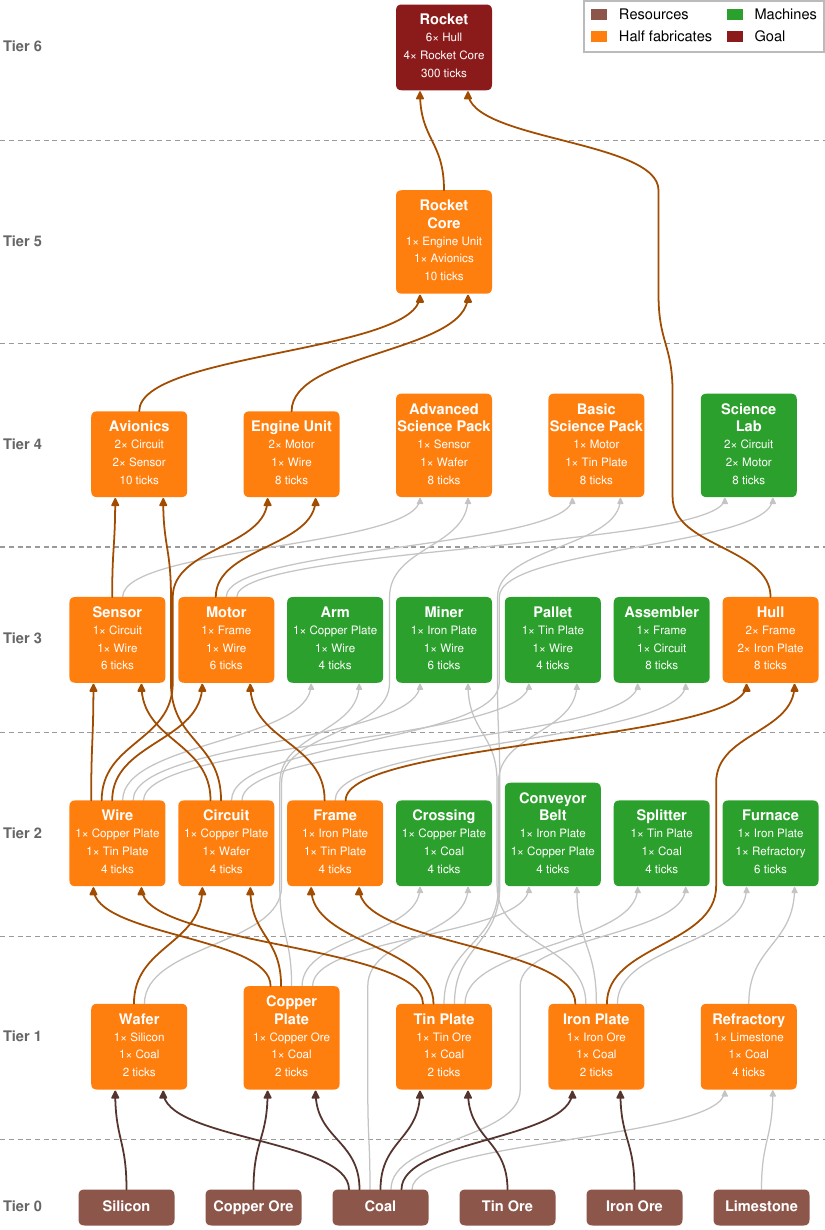}
  \caption{Standard production recipe book. Every recipe in the simulator's \texttt{BASE\_RECIPES} book, laid out by topological depth. The \easyRocket scenario uses a simplified subset (\cref{fig:recipe_dag}).}
  \label{fig:full_recipe_dag}
\end{figure}

The tables below list every item in the simulator, together with its icon, the recipe that produces it, and a short description of its function. Recipes here are the \emph{standard} recipe book (\texttt{BASE\_RECIPES}). The \easyRocket scenario described earlier in the paper swaps in a simplified set tuned to the rocket's two-stage hull/engine production line (see \cref{fig:recipe_dag}). \cref{fig:full_recipe_dag} shows the full production DAG for the standard book.

\begin{table}[ht]
  \centering
  \caption{Raw resources in the standard recipe book, with the function of each.}
  \label{tab:items_resources}
  {\small
\begin{tabular}{@{}c l p{0.55\linewidth}@{}}
\toprule
 & Name & Function \\
\midrule
\raisebox{-0.32\height}{\includegraphics[width=18pt]{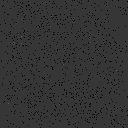}} & Coal & Mined from coal patches. Burned in furnace recipes and used in several intermediates. \\
\raisebox{-0.32\height}{\includegraphics[width=18pt]{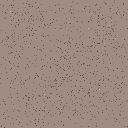}} & Iron Ore & The base metal ore. Smelts to iron plate; direct input to several machine recipes. \\
\raisebox{-0.32\height}{\includegraphics[width=18pt]{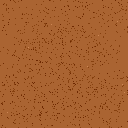}} & Copper Ore & The conductive ore. Smelts to copper plate; used in circuits and wires. \\
\raisebox{-0.32\height}{\includegraphics[width=18pt]{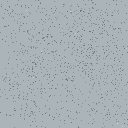}} & Tin Ore & Smelts to tin plate; used in frames, wires, and pallets. \\
\raisebox{-0.32\height}{\includegraphics[width=18pt]{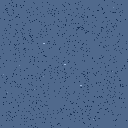}} & Silicon & Refines to wafer; input to electronic intermediates. \\
\raisebox{-0.32\height}{\includegraphics[width=18pt]{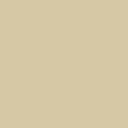}} & Limestone & Input to refractory brick and to several rocket-part recipes. \\
\bottomrule
\end{tabular}
}

\end{table}

\begin{table}[ht]
  \centering
  \caption{Machines in the standard recipe book, with the recipe that produces each one and its function.}
  \label{tab:items_machines}
  {\small
\begin{tabular}{@{}c l l p{0.36\linewidth}@{}}
\toprule
 & Name & Recipe & Function \\
\midrule
\raisebox{-0.32\height}{\includegraphics[width=18pt]{figures/icons/MINER.png}} & Miner & 1$\times$ Iron Plate, 1$\times$ Wire & Extracts ore from the tile underneath into its output buffer. \\
\raisebox{-0.32\height}{\includegraphics[width=18pt]{figures/icons/PALLET.png}} & Pallet & 1$\times$ Tin Plate, 1$\times$ Wire & Passive storage tile that holds items in place. \\
\raisebox{-0.32\height}{\includegraphics[width=18pt]{figures/icons/CONVEYOR_BELT.png}} & Conveyor Belt & 1$\times$ Iron Plate, 1$\times$ Copper Plate & Moves items one tile forward per tick along its facing direction. \\
\raisebox{-0.32\height}{\includegraphics[width=18pt]{figures/icons/ASSEMBLER.png}} & Assembler & 1$\times$ Frame, 1$\times$ Circuit & Combines two recipe inputs into an output; recipe selected by inserted items. \\
\raisebox{-0.32\height}{\includegraphics[width=18pt]{figures/icons/ARM.png}} & Arm & 1$\times$ Copper Plate, 1$\times$ Wire & Picks an item from the tile it faces and deposits it onto the opposite tile. \\
\raisebox{-0.32\height}{\includegraphics[width=18pt]{figures/icons/ROCKET.png}} & Rocket & 6$\times$ Hull, 4$\times$ Rocket Core & Capstone machine. Placing it on the map triggers liftoff. \\
\raisebox{-0.32\height}{\includegraphics[width=18pt]{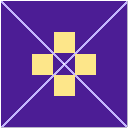}} & Science Lab & 2$\times$ Circuit, 2$\times$ Motor & Consumes science packs to progress research. \\
\raisebox{-0.32\height}{\includegraphics[width=18pt]{figures/icons/SPLITTER.png}} & Splitter & 1$\times$ Tin Plate, 1$\times$ Coal & Alternates incoming belt items between two outputs. \\
\raisebox{-0.32\height}{\includegraphics[width=18pt]{figures/icons/CROSSING.png}} & Crossing & 1$\times$ Copper Plate, 1$\times$ Coal & Two belt lanes that cross without merging. \\
\bottomrule
\end{tabular}
}

\end{table}

\begin{table}[ht]
  \centering
  \caption{Half-fabricates in the standard recipe book, with the recipe that produces each one and its function.}
  \label{tab:half_fabricates}
  {\small
\begin{tabular}{@{}c l l p{0.36\linewidth}@{}}
\toprule
 & Name & Recipe & Function \\
\midrule
\raisebox{-0.32\height}{\includegraphics[width=18pt]{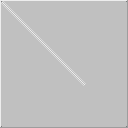}} & Iron Plate & 1$\times$ Iron Ore, 1$\times$ Coal & Smelted iron. Structural input to frames, miners, conveyors, and hulls. \\
\raisebox{-0.32\height}{\includegraphics[width=18pt]{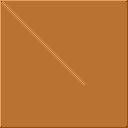}} & Copper Plate & 1$\times$ Copper Ore, 1$\times$ Coal & Smelted copper. Conductive input to circuits, wires, and conveyors. \\
\raisebox{-0.32\height}{\includegraphics[width=18pt]{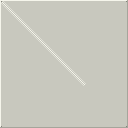}} & Tin Plate & 1$\times$ Tin Ore, 1$\times$ Coal & Smelted tin. Input to frames, wires, and pallets. \\
\raisebox{-0.32\height}{\includegraphics[width=18pt]{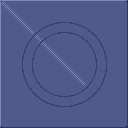}} & Wafer & 1$\times$ Silicon, 1$\times$ Coal & Refined silicon disc. Input to circuits and advanced science packs. \\
\raisebox{-0.32\height}{\includegraphics[width=18pt]{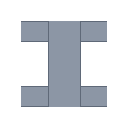}} & Frame & 1$\times$ Iron Plate, 1$\times$ Tin Plate & Structural plate-and-girder. Input to motors, assemblers, and hulls. \\
\raisebox{-0.32\height}{\includegraphics[width=18pt]{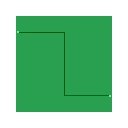}} & Circuit & 1$\times$ Copper Plate, 1$\times$ Wafer & Logic substrate. Input to sensors, assemblers, and avionics. \\
\raisebox{-0.32\height}{\includegraphics[width=18pt]{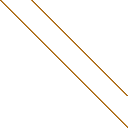}} & Wire & 1$\times$ Copper Plate, 1$\times$ Tin Plate & Conductive strand. Input to motors, sensors, miners, pallets, arms, and engine units. \\
\raisebox{-0.32\height}{\includegraphics[width=18pt]{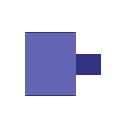}} & Motor & 1$\times$ Frame, 1$\times$ Wire & Active mechanism. Input to basic science packs and engine units. \\
\raisebox{-0.32\height}{\includegraphics[width=18pt]{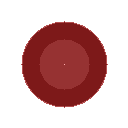}} & Sensor & 1$\times$ Circuit, 1$\times$ Wire & Detection element. Input to advanced science packs and avionics. \\
\raisebox{-0.32\height}{\includegraphics[width=18pt]{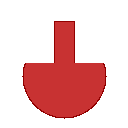}} & Basic Science Pack & 1$\times$ Motor, 1$\times$ Tin Plate & First-tier research unit. Consumed by science labs. \\
\raisebox{-0.32\height}{\includegraphics[width=18pt]{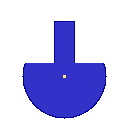}} & Advanced Science Pack & 1$\times$ Sensor, 1$\times$ Wafer & Second-tier research unit. Consumed by science labs. \\
\raisebox{-0.32\height}{\includegraphics[width=18pt]{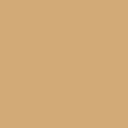}} & Refractory & 1$\times$ Limestone, 1$\times$ Coal & Heat-resistant brick. Input to furnace recipes. \\
\raisebox{-0.32\height}{\includegraphics[width=18pt]{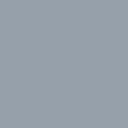}} & Hull & 2$\times$ Frame, 2$\times$ Iron Plate & Structural part of the rocket. Combines with rocket cores. \\
\raisebox{-0.32\height}{\includegraphics[width=18pt]{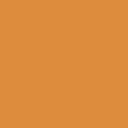}} & Engine Unit & 2$\times$ Motor, 1$\times$ Wire & Propulsion subassembly. Input to rocket cores. \\
\raisebox{-0.32\height}{\includegraphics[width=18pt]{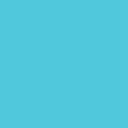}} & Avionics & 2$\times$ Circuit, 2$\times$ Sensor & Control electronics. Input to rocket cores. \\
\raisebox{-0.32\height}{\includegraphics[width=18pt]{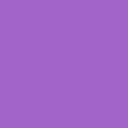}} & Rocket Core & 1$\times$ Engine Unit, 1$\times$ Avionics & Rocket subassembly. Two of these and a hull combine into a rocket. \\
\bottomrule
\end{tabular}
}

\end{table}

\clearpage

\clearpage
\section{Observation Reference}
\label{app:observation}

Factoriax ships four observation functions, formed by combining a \emph{profile} (\texttt{x\_ray} or \texttt{superficial}) with a \emph{view} (\texttt{global} or \texttt{local}), producing \texttt{global\_x\_ray}, \texttt{global\_superficial}, \texttt{local\_x\_ray}, and \texttt{local\_superficial}. All four return float32 vectors and concatenate a flattened spatial tensor with a per-player scalar block:
\[
N_\text{obs} \;=\; C_\text{prof} \cdot N_\text{tiles} \;+\; S_\text{prof}
\]
with $C_\text{prof} = 10$ for \texttt{x\_ray} and $3$ for \texttt{superficial}, $S_\text{prof} = 72$ for \texttt{x\_ray} and $63$ for \texttt{superficial}, and $N_\text{tiles}$ equal to $H \cdot W$ for the global view or $(2r+1)^2$ for the local view (default radius $r = 10$).

\subsection*{Spatial Channels}

\cref{tab:obs_spatial} lists every spatial channel and which profile includes it.

\begin{table}[ht]
  \centering
  \caption{Spatial channels per observation profile. The \texttt{x\_ray} profile uses all ten. The \texttt{superficial} profile uses three.}
  \label{tab:obs_spatial}
  {\small
\begin{tabular}{@{}r l c c l l@{}}
\toprule
\# & Channel & x\_ray & superficial & Encodes & Normalisation \\
\midrule
1 & \texttt{block\_type} & \checkmark & \checkmark & Tile type (dirt, water, ore type, ...) & $/\max(\texttt{BlockType})$ \\
2 & \texttt{machine\_type} & \checkmark & \checkmark & Machine type placed on the tile (or NONE) & $/\max(\texttt{Machine})$ \\
3 & \texttt{block\_resources} & \checkmark & --- & Ore count remaining under the tile & $/30000$ \\
4 & \texttt{slot0\_type} & \checkmark & --- & Assembler input slot 0 item type & $/33$ \\
5 & \texttt{slot0\_count} & \checkmark & --- & Assembler input slot 0 item count & $/1024$ \\
6 & \texttt{slot1\_type} & \checkmark & --- & Assembler input slot 1 item type & $/33$ \\
7 & \texttt{slot1\_count} & \checkmark & --- & Assembler input slot 1 item count & $/1024$ \\
8 & \texttt{slot2\_type} & \checkmark & --- & Assembler output / buffer slot item type & $/33$ \\
9 & \texttt{slot2\_count} & \checkmark & --- & Assembler output / buffer slot item count & $/1024$ \\
10 & \texttt{machine\_direction} & \checkmark & \checkmark & Cardinal facing of the placed machine & $/4$ \\
\bottomrule
\end{tabular}
}

\end{table}

\subsection*{Per-Player Scalars}

Both profiles share a common scalar block. The \texttt{x\_ray} profile additionally appends nine facing-tile scalars covering the machine state of the tile directly in front of the player.

\begin{table}[ht]
  \centering
  \caption{Common per-player scalars, shared by both profiles. Recipe-affordability and inventory sizes scale with the active recipe book and item set.}
  \label{tab:obs_common}
  {\small
\begin{tabular}{@{}l c p{0.55\linewidth}@{}}
\toprule
Group & Count & Encodes \\
\midrule
Pose + time & 4 & Player $x$ (normalised by \texttt{map\_width}), $y$ (by \texttt{map\_height}), facing direction ($/4$), current timestep (by \texttt{max\_timesteps}) \\
Recipe affordability & \texttt{NUM\_RECIPES} (26 in the default book) & One float per recipe in the active book; 1 if the player can hand-craft the recipe right now, 0 otherwise \\
Player inventory & \texttt{NUM\_ITEM\_TYPES} (33) & Item count per type, normalised by per-item \texttt{PLAYER\_MAX\_STACK} \\
\midrule
Total (both profiles) & 63 & \\
\bottomrule
\end{tabular}
}

\end{table}

\begin{table}[ht]
  \centering
  \caption{Facing-tile scalars appended by the \texttt{x\_ray} profile only. Each scalar reads the entity on the tile directly in front of the player. All nine are zero if that tile is out of bounds.}
  \label{tab:obs_facing}
  {\small
\begin{tabular}{@{}r l p{0.55\linewidth}@{}}
\toprule
\# & Scalar & Encodes \\
\midrule
1 & \texttt{facing\_machine\_type} & Machine type on the tile in front of the player \\
2 & \texttt{facing\_buffer\_type} & Buffer item type on that tile \\
3 & \texttt{facing\_buffer\_count} & Buffer item count on that tile \\
4 & \texttt{facing\_asm\_in0\_type} & Assembler input slot 0 item type \\
5 & \texttt{facing\_asm\_in0\_count} & Assembler input slot 0 item count \\
6 & \texttt{facing\_asm\_in1\_type} & Assembler input slot 1 item type \\
7 & \texttt{facing\_asm\_in1\_count} & Assembler input slot 1 item count \\
8 & \texttt{facing\_asm\_out\_type} & Assembler output item type \\
9 & \texttt{facing\_asm\_out\_count} & Assembler output item count \\
\bottomrule
\end{tabular}
}

\end{table}

\newpage
\section{Action Reference}
\label{app:action}


The action space is a single discrete head of size $\texttt{NUM\_ACTIONS} = 85$, laid out as five categories ordered \texttt{MoveAction}, \texttt{InteractAction}, \texttt{PLACE\_*}, \texttt{CRAFT\_*}, and \texttt{DEPOSIT\_*}. The two non-parametric families (\cref{tab:action_movement,tab:action_interact}) name every action explicitly. The three parametric families~(\cref{tab:action_parametric}) are summarised by base offset, count, and the item set that they describe~(\cref{app:items}).

\begin{table}[ht]
  \centering
  \caption{Movement family. Codes are the absolute \texttt{Action} integer (movement starts at 0). Cardinal moves implicitly set facing. \texttt{FACE\_*} actions snap facing without moving.}
  \label{tab:action_movement}
  {\small
\begin{tabular}{@{}r l p{0.55\linewidth}@{}}
\toprule
Code & Action & Effect \\
\midrule
0 & \texttt{NOOP} & Skip the tick; player state is unchanged \\
1 & \texttt{UP} & Move one tile up if passable, otherwise stay; sets facing UP \\
2 & \texttt{DOWN} & Move one tile down if passable, otherwise stay; sets facing DOWN \\
3 & \texttt{LEFT} & Move one tile left if passable, otherwise stay; sets facing LEFT \\
4 & \texttt{RIGHT} & Move one tile right if passable, otherwise stay; sets facing RIGHT \\
5 & \texttt{FACE\_UP} & Snap facing direction to UP without moving \\
6 & \texttt{FACE\_DOWN} & Snap facing direction to DOWN without moving \\
7 & \texttt{FACE\_LEFT} & Snap facing direction to LEFT without moving \\
8 & \texttt{FACE\_RIGHT} & Snap facing direction to RIGHT without moving \\
\bottomrule
\end{tabular}
}

\end{table}

\begin{table}[ht]
  \centering
  \caption{Interaction family. Codes follow movement (starting at 9). Each action operates on the tile directly in front of the player.}
  \label{tab:action_interact}
  {\small
\begin{tabular}{@{}r l p{0.55\linewidth}@{}}
\toprule
Code & Action & Effect \\
\midrule
9 & \texttt{MINE} & Reduce the ore counter on the tile in front by one and add the ore to the player inventory \\
10 & \texttt{PICKUP} & Pick up the machine on the tile in front into the player inventory \\
11 & \texttt{WITHDRAW} & Drain the output slot of the machine in front into the player inventory \\
12 & \texttt{REPAIR} & Restore the entity in front to full health (see note below) \\
13 & \texttt{ROTATE\_LEFT} & Rotate the machine on the tile in front to face LEFT \\
14 & \texttt{ROTATE\_RIGHT} & Rotate the machine on the tile in front to face RIGHT \\
15 & \texttt{ROTATE\_UP} & Rotate the machine on the tile in front to face UP \\
16 & \texttt{ROTATE\_DOWN} & Rotate the machine on the tile in front to face DOWN \\
\bottomrule
\end{tabular}
}

\end{table}

The base simulator implements \texttt{REPAIR} as a full health restore on the entity in front of the player, but the simulator currently ships without a damage mechanic, so the handler has no observable effect at present. Damage propagation and the matching repair-cost rules are future work.

\begin{table}[ht]
  \centering
  \caption{Parametric families. Each family covers a contiguous range of \texttt{Action} integers starting at the listed base offset. The item index inside the family follows the simulator's enum order.}
  \label{tab:action_parametric}
  {\small
\begin{tabular}{@{}l r r p{0.50\linewidth}@{}}
\toprule
Family & Base offset & Count & Items \\
\midrule
\texttt{PLACE\_*} & 17 & 10 & \texttt{Machine} (NONE excluded); see Appendix~\ref{app:items}, Table~\ref{tab:machines} \\
\texttt{CRAFT\_*} & 27 & 26 & \texttt{HalfFabricate} $\cup$ \texttt{Machine} (NONE excluded); see Appendix~\ref{app:items}, Tables~\ref{tab:half_fabricates} and \ref{tab:machines} \\
\texttt{DEPOSIT\_*} & 53 & 32 & \texttt{Resource} $\cup$ \texttt{HalfFabricate} $\cup$ \texttt{Machine} (NONE excluded); see Appendix~\ref{app:items} \\
\bottomrule
\end{tabular}
}

\end{table}

\newpage
\section{Reproducibility}
\label{app:reproducibility}


\subsection*{Seeds}
Seeds reported in this paper. \textbf{Seed 42} for the scripted-baseline build progression (\cref{fig:build_progression}) and the final-factory heatmap (\cref{fig:build_heatmap}). \textbf{Seeds 0--3} for the reset-grid sample (\cref{fig:reset_grid}). \textbf{Seeds 0--15} for the scripted-baseline timing distribution (\cref{fig:scripted_timeline}). PPO training runs were run with \textbf{Seeds 0-9}. PPO throughput numbers (\cref{fig:throughput}) are single runs per (device, batch size, map size, entity cap) cell, with the wall-clock time measured directly. Throughput numbers carry no seed dependence beyond JAX's PRNG init for the env reset.

\subsection*{Hardware}
Numbers reported in this paper were measured on two devices: an NVIDIA GeForce RTX 3050 6GB laptop GPU and an NVIDIA A100. Both run JAX 0.9.1 against the CUDA backend with no custom XLA flags set. The throughput sweep (\cref{fig:throughput}) takes one timed run per (device, batch size, map size, entity cap) cell, preceded by a JIT warm-up iteration that is excluded from the measurement.

\subsection*{PPO Hyperparameters}
\begin{tabular}{@{}lll@{}}
\toprule
\textbf{Hyperparameter} & \textbf{10-billion-step run} & \textbf{Trainer default} \\
\midrule
Learning rate                  & $1 \times 10^{-4}$        & $2.5 \times 10^{-4}$ \\
Entropy coefficient            & $0.01$                    & $0.01$ \\
Discount factor $\gamma$       & $0.99$                    & $0.99$ \\
GAE smoothing $\lambda$        & $0.95$                    & $0.95$ \\
Clip $\epsilon$                & $0.2$                     & $0.2$ \\
Value-loss coefficient         & $0.5$                     & $0.5$ \\
Update epochs per iteration    & $4$                       & $4$ \\
Minibatches per epoch          & $8$                       & $8$ \\
Max gradient norm              & $0.5$                     & $0.5$ \\
Parallel environments          & $4096$                    & $64$ \\
Rollout length (steps)         & $128$                     & $128$ \\
Network hidden dimensions      & $(256, 256)$              & $(256, 256)$ \\
Observation normalisation      & online running mean/std   & online running mean/std \\
Total steps                    & $10^{10}$                 & $10^{7}$ \\
Optimiser                      & Adam, linear LR anneal    & Adam, linear LR anneal \\
\bottomrule
\end{tabular}
\vspace{0.4em}

The 10-billion-step run column lists the values used for the 10-seed PPO results (\cref{sec:experiments}). Network architecture is a flat MLP shared between the policy and value heads, with the heads as the final two linear layers.

\newpage
\section{Easy Rocket Scenario Details}
\label{app:easy_rocket}

\begin{figure}[ht]
  \centering
  \includegraphics[width=0.7\linewidth]{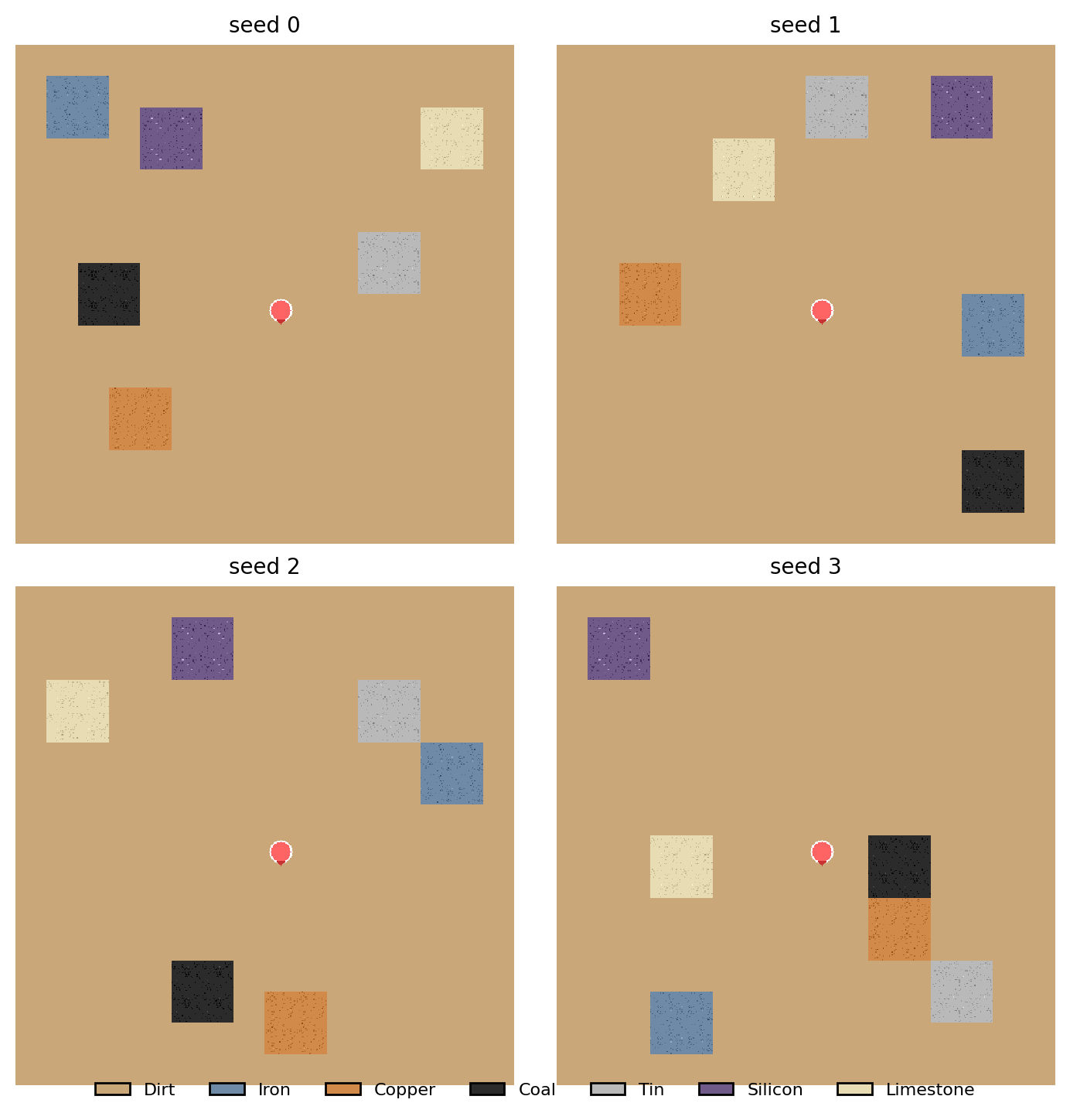}
  \caption{Four reset states of \easyRocket. The player spawns at the centre of a 16$\times$16 grid scattered with patches of finite raw materials. The patches are placed differently for each random seed while preserving spawn-area clearance and a perimeter around the ore patches for belt routing.}
  \label{fig:reset_grid}
\end{figure}

\begin{figure}[H]
  \centering
  \includegraphics[width=0.7\linewidth]{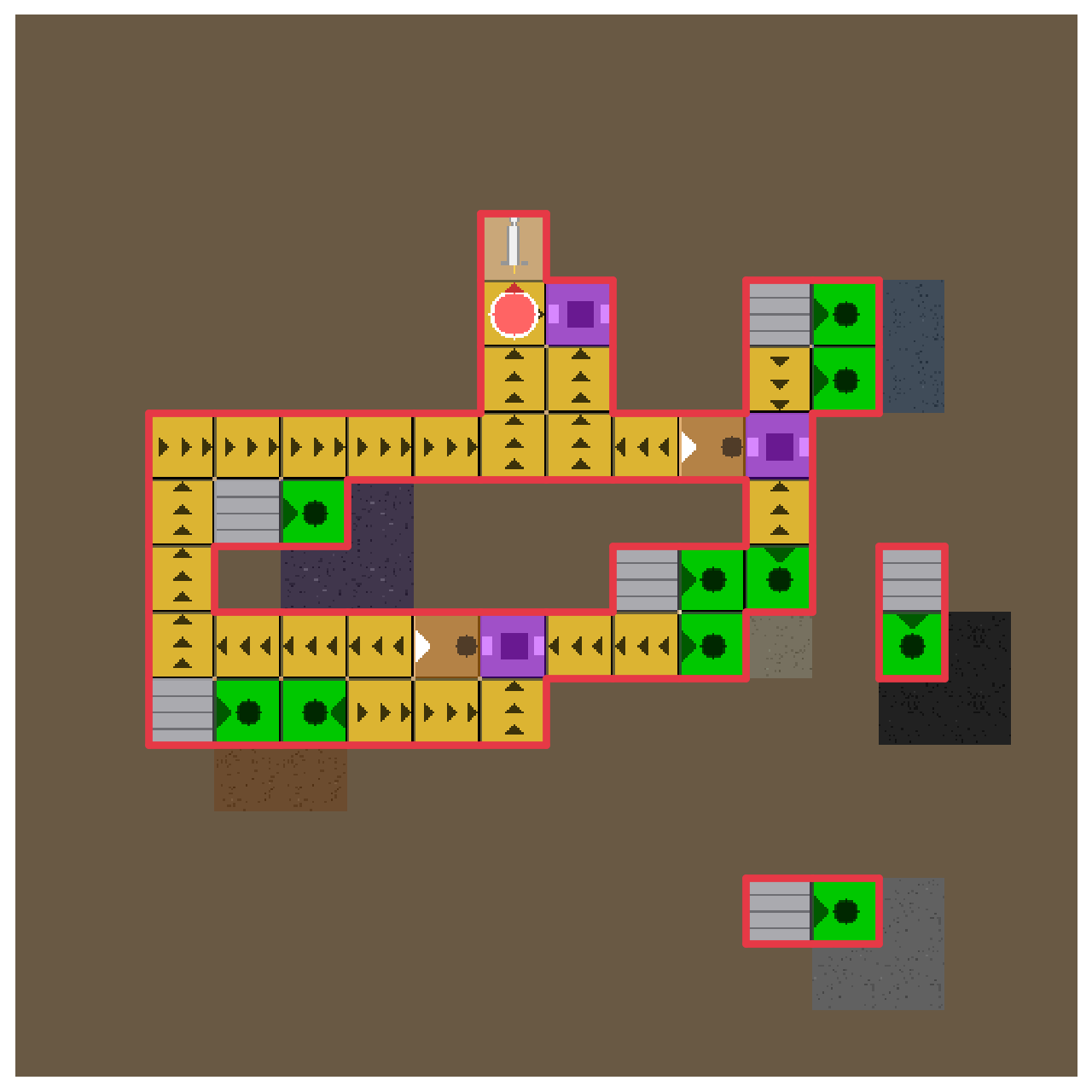}
  \caption{Final scripted-baseline factory on seed 42 for \easyRocket. Six independent miner setups, four more for the production lines. Three assemblers, two of which produce hulls and engines, and the final one combining the two to create the rocket according the the recipe graph in Figure \ref{fig:recipe_dag}.}
  \label{fig:build_heatmap}
\end{figure}

\cref{sec:benchmark} describes the \easyRocket scenario. This appendix goes into more detail. \cref{fig:reset_grid} shows four random initialisations of the map. \cref{tab:curriculum} describes each achievement in the benchmark. \cref{fig:build_heatmap} shows the factory created by the scripted baseline on seed 42.

\begin{table}[ht]
  \centering
  \caption{\easyRocket achievements in unlock order. Achievements 1-3 read the state of the  agent inventory, and achievements 4-25 read the map state for placed machines or item buffer states.}
  \label{tab:curriculum}
  {\small
\begin{tabular}{@{}rllp{0.42\linewidth}@{}}
\toprule
\# & Engine name & Phase & Triggered when \\
\midrule
\rowcolor[HTML]{eff5fb} 1 & \texttt{mine\_1\_ore} & Bootstrap & Player inventory holds any raw ore. \\
\rowcolor[HTML]{eff5fb} 2 & \texttt{mine\_3\_ore} & Bootstrap & Player inventory holds at least one of three distinct raw ore types. \\
\rowcolor[HTML]{eff5fb} 3 & \texttt{prospector} & Bootstrap & Player inventory holds at least one of every raw ore type. \\
\rowcolor[HTML]{eff5fb} 4 & \texttt{place\_miner} & Bootstrap & At least one miner is placed on the map. \\
\rowcolor[HTML]{eff5fb} 5 & \texttt{place\_3\_miners} & Bootstrap & Three miners are placed on the map. \\
\rowcolor[HTML]{eff5fb} 6 & \texttt{metal\_in\_motion} & Bootstrap & Six miners are placed on the map. \\
\midrule
\rowcolor[HTML]{d4e3f3} 7 & \texttt{automated\_mining} & Miners up & A placed miner has ore in its output buffer. \\
\rowcolor[HTML]{d4e3f3} 8 & \texttt{ore\_fields} & Miners up & Three or more ore types sit under producing miners. \\
\rowcolor[HTML]{d4e3f3} 9 & \texttt{mining\_master} & Miners up & All six ore types sit under producing miners. \\
\midrule
\rowcolor[HTML]{bcd2eb} 10 & \texttt{one\_pallet} & Storage & At least one pallet is placed on the map. \\
\rowcolor[HTML]{bcd2eb} 11 & \texttt{three\_pallets} & Storage & At least three pallets are placed on the map. \\
\rowcolor[HTML]{bcd2eb} 12 & \texttt{stacked} & Storage & At least six pallets are placed on the map. \\
\midrule
\rowcolor[HTML]{9fc1df} 13 & \texttt{assembler\_online} & Hull & At least one assembler is placed on the map. \\
\rowcolor[HTML]{9fc1df} 14 & \texttt{assembler\_and\_arm} & Hull & At least one assembler and one arm are placed on the map. \\
\rowcolor[HTML]{9fc1df} 15 & \texttt{one\_belt} & Hull & At least one conveyor belt is placed on the map. \\
\rowcolor[HTML]{9fc1df} 16 & \texttt{three\_belts} & Hull & At least three conveyor belts are placed on the map. \\
\rowcolor[HTML]{9fc1df} 17 & \texttt{hull\_production} & Hull & An assembler carries a finished hull in output or buffer. \\
\midrule
\rowcolor[HTML]{7aabd2} 18 & \texttt{two\_assemblers} & Engine & A second assembler is placed on the map, opening a parallel production line. \\
\rowcolor[HTML]{7aabd2} 19 & \texttt{an\_arm\_and\_a\_leg} & Engine & At least two assemblers and two arms are placed on the map. \\
\rowcolor[HTML]{7aabd2} 20 & \texttt{six\_belts} & Engine & At least six conveyor belts are placed on the map. \\
\rowcolor[HTML]{7aabd2} 21 & \texttt{engine\_production} & Engine & An assembler carries a finished engine unit. \\
\midrule
\rowcolor[HTML]{4a87bb} 22 & \texttt{auto\_bots} & Rocket production & A third assembler is placed on the map, freeing one for the rocket-assembly line. \\
\rowcolor[HTML]{4a87bb} 23 & \texttt{belt\_spaghetti} & Rocket production & At least ten conveyor belts are placed on the map. \\
\rowcolor[HTML]{4a87bb} 24 & \texttt{rocket\_assembled} & Rocket production & An assembler carries a finished rocket item. \\
\rowcolor[HTML]{4a87bb} 25 & \texttt{liftoff} & Rocket production & A rocket machine has been placed on the map. \\
\bottomrule
\end{tabular}
}

\end{table}

\newpage
\newpage
\section{PPO Training}
\label{app:param_variation_grid}

Per-phase breakdown of the ten-seed run from \cref{sec:experiments}. Each row in the graph corresponds to a phase in the achievement set.
Each phase cell shows the individual achievement unlock rates shown in figure (\cref{fig:achievement_buckets}).

\begin{figure}[ht]
  \centering
  \includegraphics[width=0.75\linewidth, keepaspectratio=true]{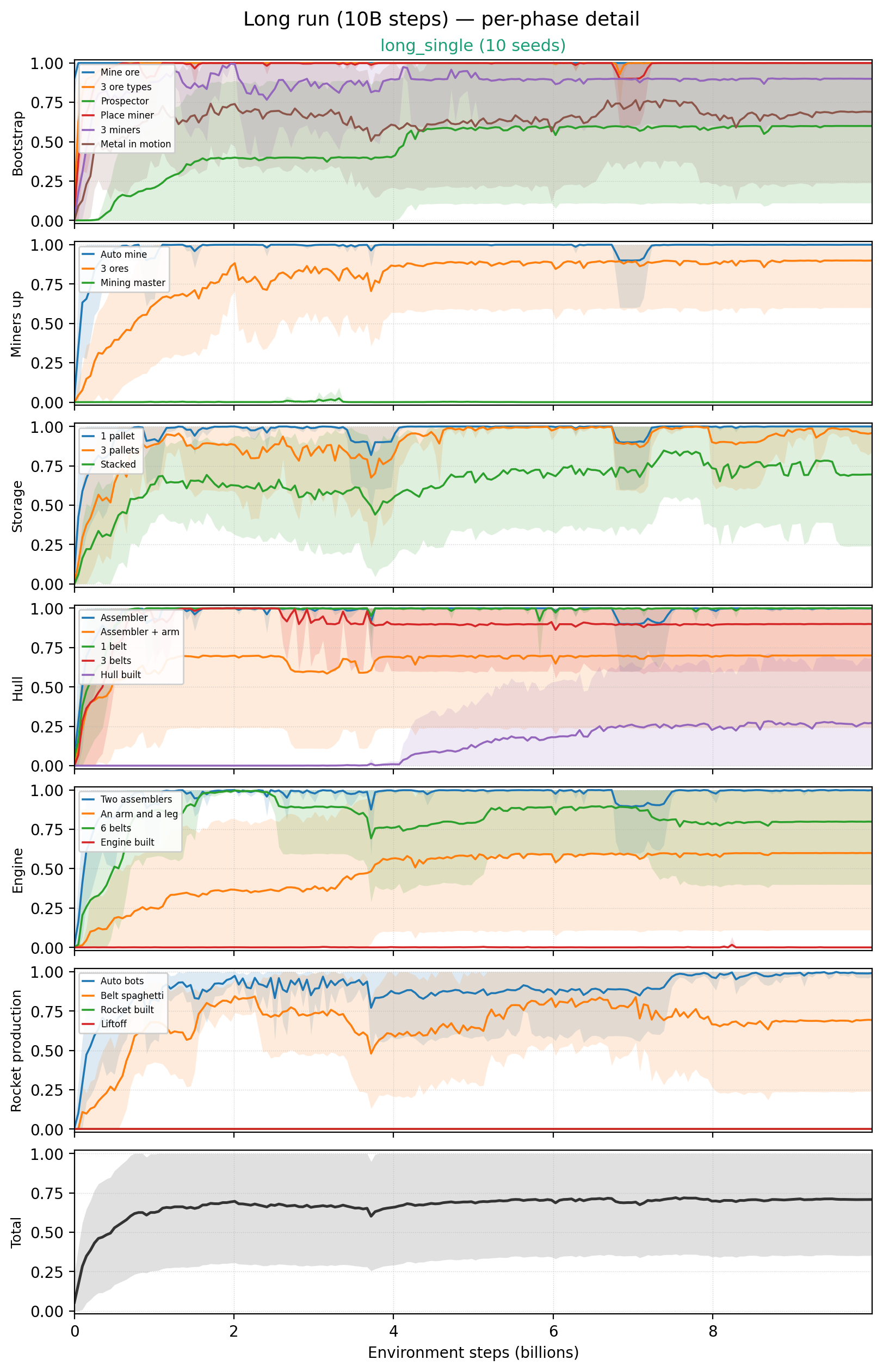}
  \caption{10-billion-step run training runs. Each plot shows achievement unlock for that phase. Rates reported over training (ten seeds).}
  \label{fig:achievement_grid_long_run}
\end{figure}

\end{document}